\documentclass{article}

\usepackage{arxiv}
\usepackage[utf8]{inputenc} 
\usepackage[T1]{fontenc}    
\usepackage{hyperref}       
\usepackage{url}            
\usepackage{booktabs}       
\usepackage{amsfonts}       
\usepackage{nicefrac}       
\usepackage{microtype}      
\usepackage{lipsum}		    
\usepackage{graphicx}
\usepackage{natbib}
\usepackage{doi}
\usepackage{xcolor}
\usepackage{siunitx}
\usepackage{enumitem}

\hypersetup{
    colorlinks,
    linkcolor={red!50!black},
    citecolor={blue!50!black},
    urlcolor={blue!80!black}
}

\title{Protecting President Zelenskyy against Deep Fakes}

\author{Maty{\'a}{\v{s}} Boh{\'a}{\v{c}}ek \\
    Gymnasium of Johannes Kepler \\
    Parléřova 2/118, 169 00 Praha 6, Czech Republic \\
	\texttt{matyas.bohacek@matsworld.io} \\
	\And
	Hany Farid \\
	Department of Electrical Engineering and Computer Sciences \\
	School of Information \\
	University of California, Berkeley \\
	\texttt{hfarid@berkeley.edu}\\
}

\date{}

\begin{document}
\maketitle

\begin{abstract}
	The 2022 Russian invasion of Ukraine is being fought on two fronts: a brutal ground war and a duplicitous disinformation campaign designed to conceal and justify Russia's actions. This campaign includes at least one example of a deep-fake video purportedly showing Ukrainian President Zelenskyy admitting defeat and surrendering. In anticipation of future attacks of this form, we describe a facial and gestural behavioral model that captures distinctive characteristics of Zelenskyy's speaking style. Trained on over eight hours of authentic video from four different settings, we show that this behavioral model can distinguish Zelenskyy from deep-fake imposters.This model can play an important role -- particularly during the fog of war -- in distinguishing the real from the fake.
\end{abstract}

\keywords{Deep fakes \and Disinformation \and Digital Forensics \and Facial Mannerisms \and Gestural Mannerisms}

\section{Introduction}
\label{sec:introduction}

\begin{figure}[b!]
    \begin{center}
    \begin{tabular}{cc}
        \includegraphics[height=4.5cm]{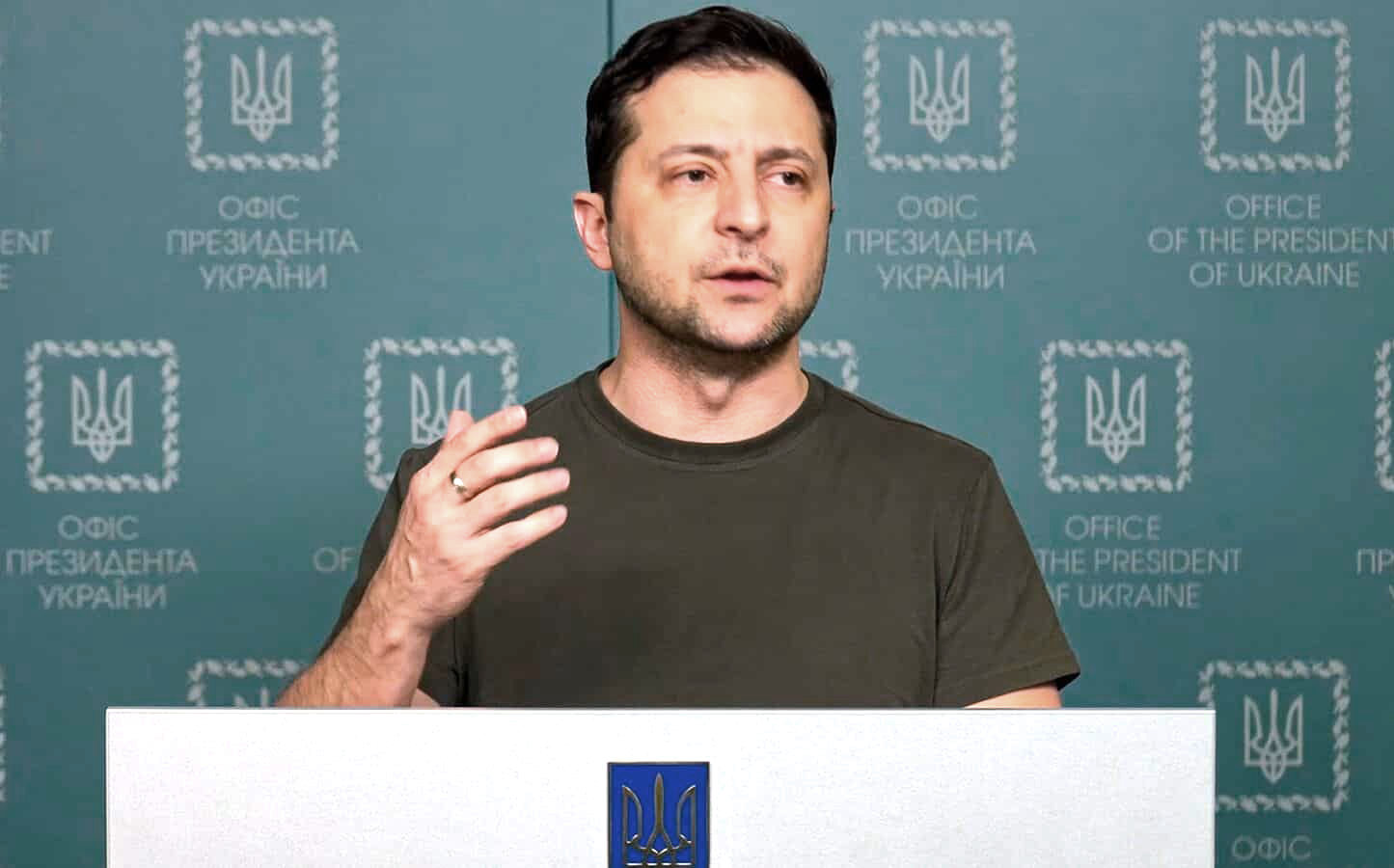} &
        \includegraphics[height=4.5cm]{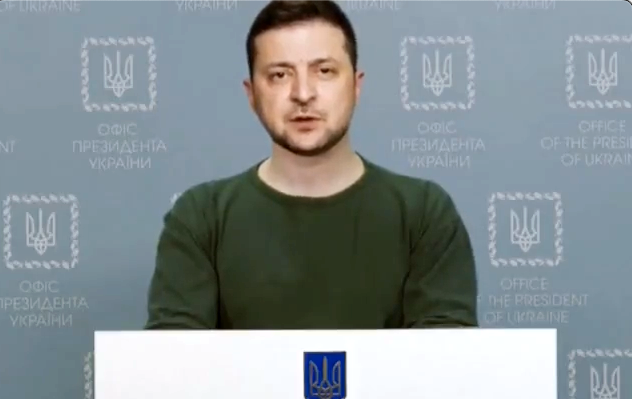}
        \end{tabular}
    \end{center}
    \caption{One video frame of the real (left) and deep-fake version (right) of Ukrainian President Zelenskyy.}
    \label{fig:zelenskyy-real-fake}
\end{figure}

In the early days of the Russian invasion of Ukraine, President Zelenskyy warned the world that Russia's digital disinformation machinery would create a deep fake of him admitting defeat and surrendering. A few weeks later in mid-March of 2022, a deep fake of Zelenskyy appeared with just this message~\citep{allyn2022deepfake}. This video, Figure~\ref{fig:zelenskyy-real-fake}, was quickly debunked thanks to the rather crude audio and video and to Zelenskyy's pre-bunking. This type of deep fake, however, is likely just the beginning of a new digital front that we might expect in this and future conflicts.

A recent set of perceptual studies~\citep{groh2022deepfake} examined the ability of untrained observers to distinguish between real and deep-fake videos. In one condition, participants viewed a single video and categorized it as real or fake. Participants correctly identified ~$66\%$ of the deep-fake videos, as compared to chance performance of $50\%$ (pooled responses from all participants -- so-called crowd wisdom -- yields an improved accuracy of $80\%$). In a second condition, participants were shown the prediction by the top-performing DFDC computational model~\citep{dfdc2020} and given the opportunity to update their response. In this collaborative condition, individual participant accuracy improved to $73\%$. 

While we may have the ability to perceptually detect some deep-fake videos, our ability is not terribly reliable and this task will become increasingly more difficult as deep fakes continue to improve in quality and sophistication. We must, therefore, turn to computational methods to assist in the task of distinguishing the real from the fake.

The computational detection of deep-fake videos can be partitioned into three basic categories: (1) learning-based, in which features that distinguish real from fake content are explicitly learned by any of a range of different machine-learning techniques~\citep{zhou2017twostream,afchar2018mesonet,li2019xray}; (2) artifact-based, in which a range of low-level (pixel based) to high-level (semantic based) features are explicitly designed to distinguish between real and fake content~\citep{li2018ictu,agarwal2021detecting,agarwal2020phoneme}; and (3) identity-based, in which biometric-style features are used to identify if the person depicted in a video is who it purports to be~\citep{agarwal2019protecting,agarwal2020appearance,agarwal2021watch,cozzolino2021id}. 

The advantage of learning-based approaches is they are able to learn detailed and subtle video-synthesis artifacts. The disadvantage is these techniques often struggle to generalize to new content not explicitly part of the training data set, and can be vulnerable to adversarial attacks~\citep{carlini2020evading}, and simple laundering attacks where the synthesized media is trans-coded or resized~\citep{barni2018adversarial}. In the 2019-2020 Deepfake Detection Challenge~\citep{dfdc2020}, for example, $2116$ teams competed for one million dollars (USD) in prizes. Teams were provided $23,654$ real videos and $104,500$ deep-fake videos created from the provided real videos. The top performing learning-based detector achieved a detection accuracy of only $65\%$ on a set of $4000$ holdout videos, half of which were real and half of which were deep fakes (i.e.,~chance performance is $50\%$). These results reveal that fully automatic detection of deep fakes in the wild remains a challenging problem. 

On the other hand, the advantage of artifact-based techniques is they can exploit inconsistencies that are difficult to circumvent or launder. The disadvantage is these techniques are typically narrowly applicable to a subset of deep-fake videos, and often require human annotation as part of the process.

The advantage of identity-based techniques is they are also resilient to adversarial and laundering attacks and are typically applicable to many different forms of deep fakes. The disadvantage of these approaches is an identity-specific model must be constructed for each individual, typically from hours of authentic video footage. This may be practical when it comes to, for example, protecting a few world leaders from deep fakes -- for which hours of video can typically be found online -- but is otherwise impractical. The other disadvantage is that the learned mannerisms are somewhat context dependent: when a world leader is giving a public address, for example, she may be more formal than when she is giving an unscripted interview, and so the specific mannerisms may not generalize across different contexts.

Because we are focused here on protecting one world leader -- Ukrainian President Zelenskyy -- and because we can easily acquire hours of video of Zelenskyy, we contend that an identity-based approach is the most sensible and robust approach. We start with the identity-based technique of~\citep{agarwal2019protecting}, leveraging distinct patterns of facial and head movements, to distinguish Zelenskyy from an imposter or deep fake. We then augment this identity-based model with new gestural features capturing how a speaker uses their hands when speaking.

After reviewing the facial mannerisms portion of the model and describing the new gestural mannerisms portion, we evaluate the efficacy of our model in distinguishing Zelenskyy from deep-fake Zelenskyy and a range of other identities.

\begin{figure}[t]
    \begin{center}
    \begin{tabular}{cccc}
    (a) & (b) & (c) & (d) \\
        \includegraphics[height=3.5cm]{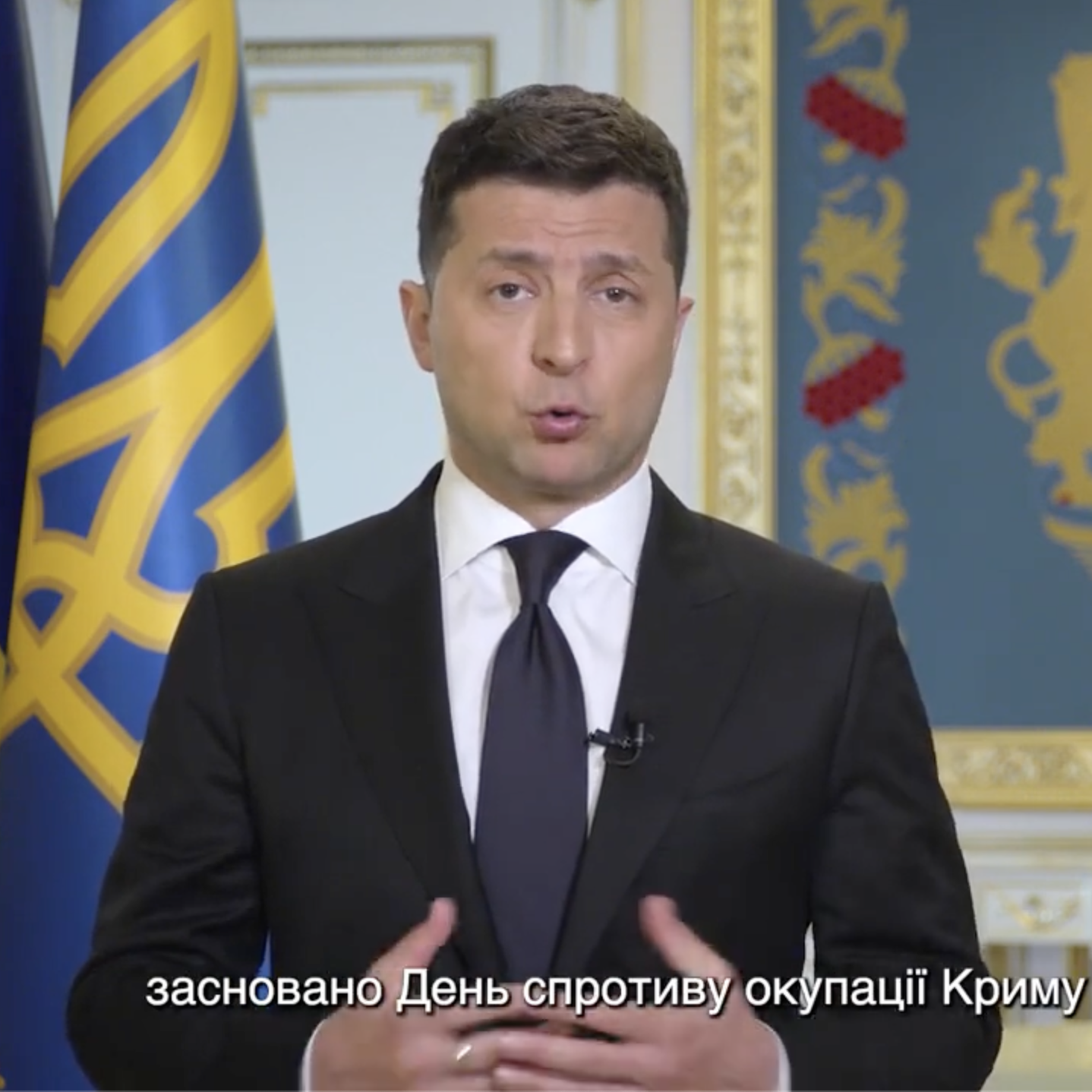} &
        \includegraphics[height=3.5cm]{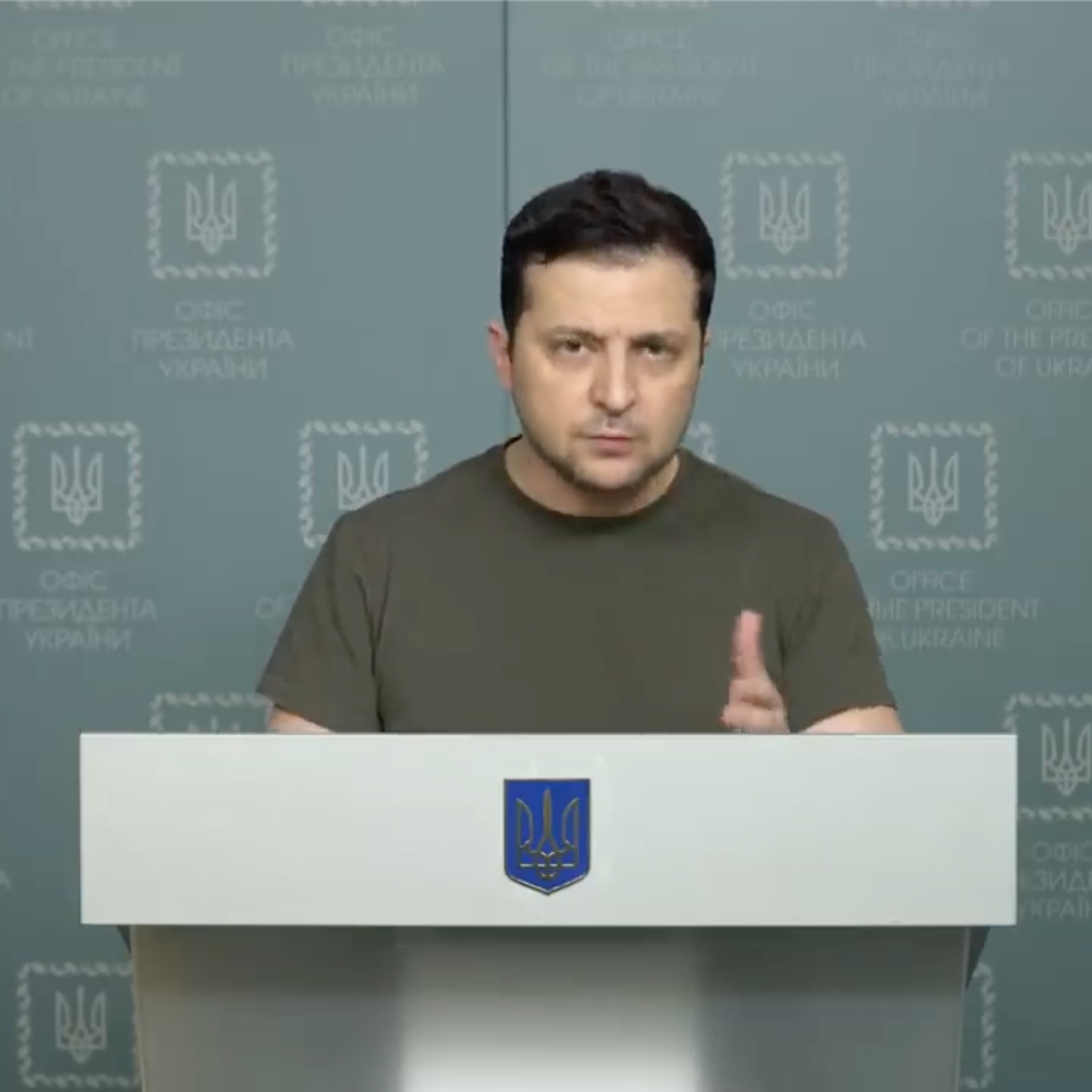} &
        \includegraphics[height=3.5cm]{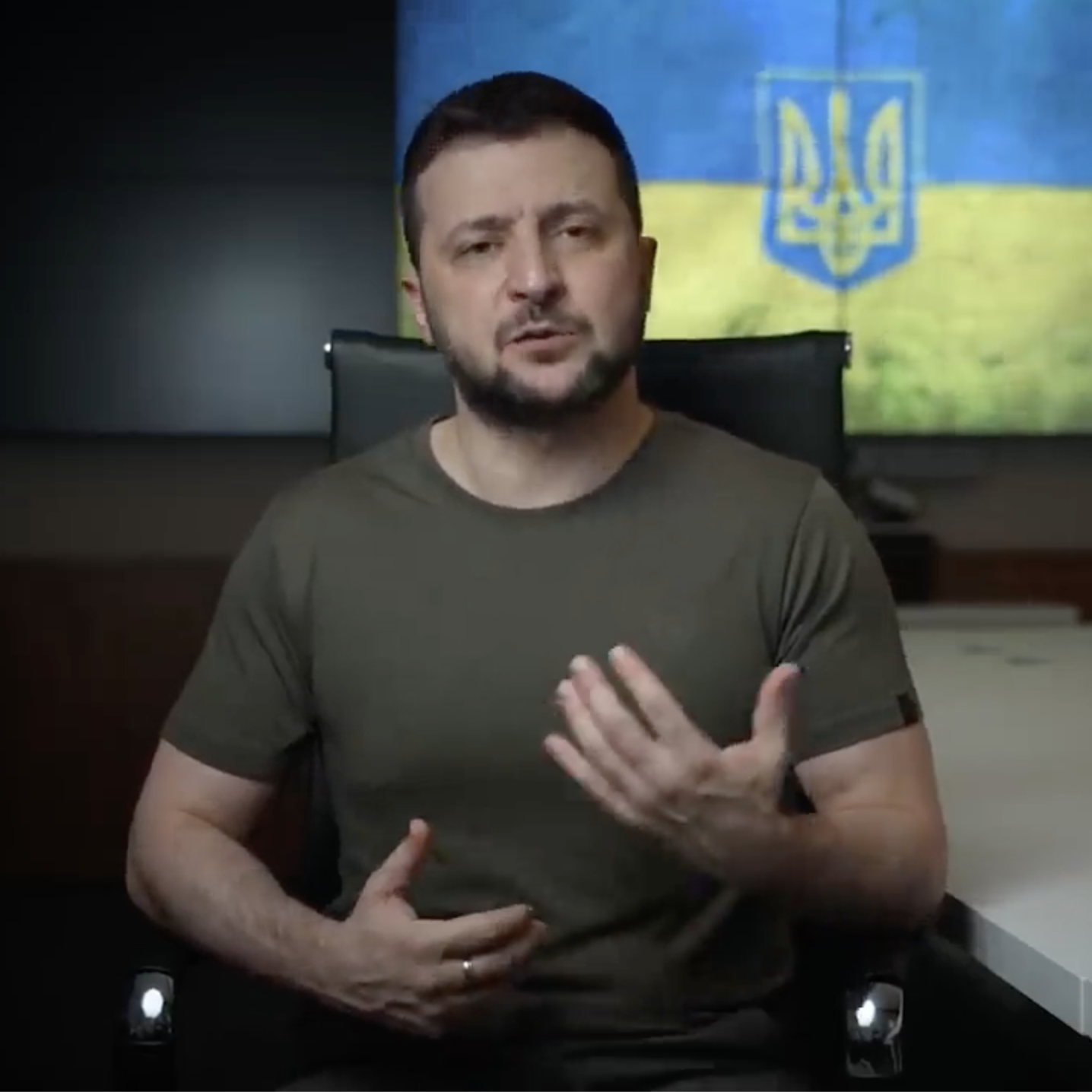} &
        \includegraphics[height=3.5cm]{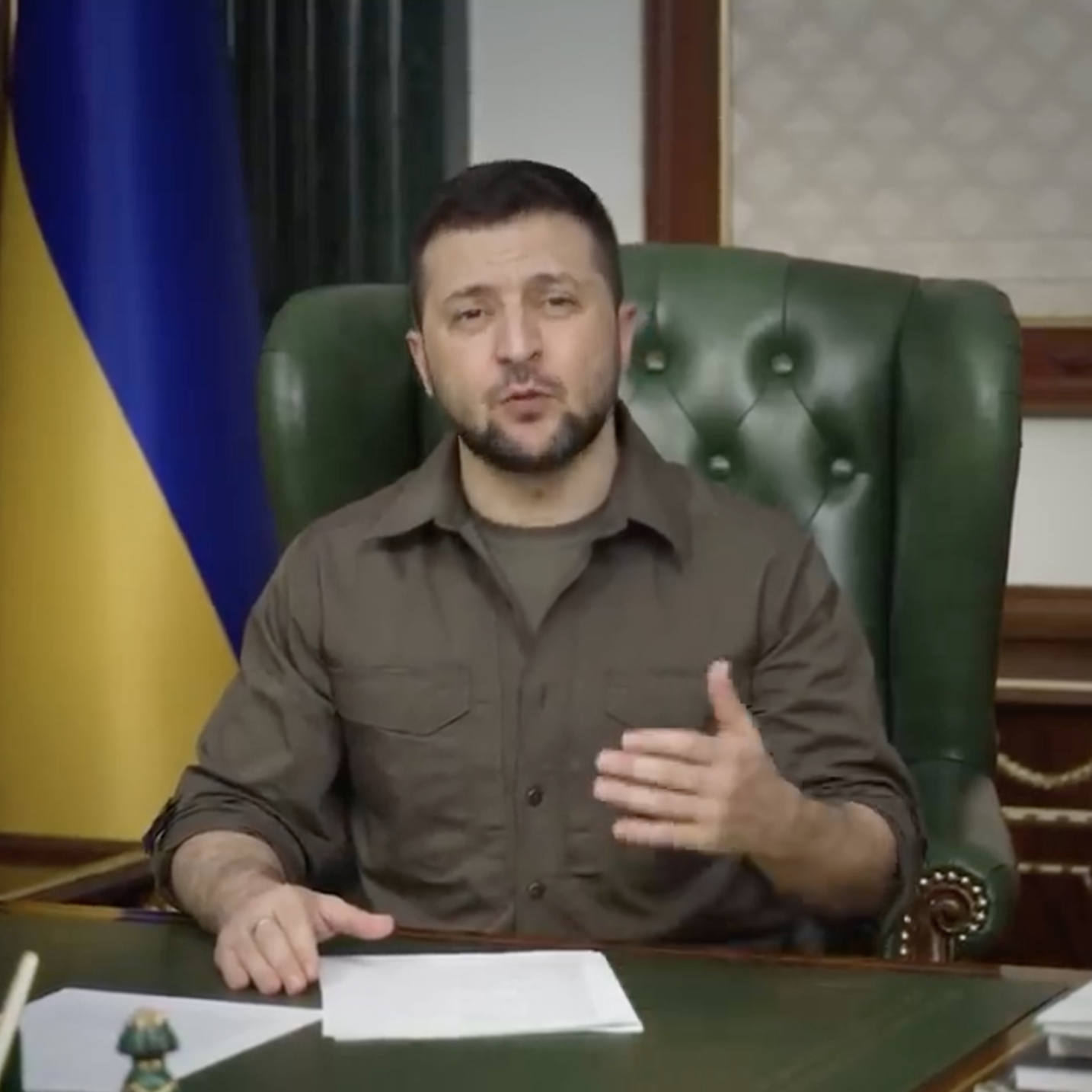} 
    \end{tabular}
    \end{center}
    \caption{Four representative examples of President Zelenskyy in different contexts: (a) public address; (b) press briefing; (c) bunker; and (d) armchair.}
    \label{fig:context}
\end{figure}
%
%

\section{Methods}
\label{sec:methods}

\subsection{Data Set}
\label{subsec:dataset}

We downloaded $506$ minutes of video of Zelenskyy from YouTube and the official website of the office of the Ukranian president\footnote{\url{https://www.president.gov.ua/en/videos/videos-archive}} in four different contexts: (a) public address ($91$ min); (b) press briefing ($207$ min); (c) bunker ($47$ min); and (d) armchair ($161$ min). Shown in Figure~\ref{fig:context} are representative examples from each of these settings.

Portions of each video with large camera motions (e.g.,~zoom, translation, cross-fade) were automatically detected and removed from the data set. In particular, the inter-frame difference was computed between each successive pair of video frames. Assuming each video depicts a speaker in the center of the frame, a camera motion was detected if the absolute difference on the left and right margin (defined as $10\%$ of the frame width) was above a specified threshold. 

A total of $57$ minutes of interview-style videos of seven world leaders (Jacinda Ardern, Joe Biden, Kamala Harris, Boris Johnson,  Wladimir Klitschko, Angela Merkel, and Vladimir Putin) were used as decoys (i.e.,~not Zelenskyy). Our deep-fake detection is designed to distinguish Zelenskyy's behavioral and gestural mannerisms from imposters driving the creation of a deep fake, and so these decoy videos -- regardless of the identities -- serve as proxies for deep fakes. An additional $50$ minutes of video across $27$ distinct individuals taken from the FaceForensics++~\citep{roessler2019faceforensicspp} dataset were used as additional decoys. In addition to these proxies, three commissioned lip-sync deep fakes ($2$ min) created by the team at Colossyan\footnote{\url{https://www.colossyan.com}}, and one in-the-wild deep fake ($1$ min) were added to this decoy dataset (Figure~\ref{fig:zelenskyy-real-fake}).

\begin{figure}[t]
    \begin{center}
    \begin{tabular}{ccccc}
        \includegraphics[width=0.165\linewidth]{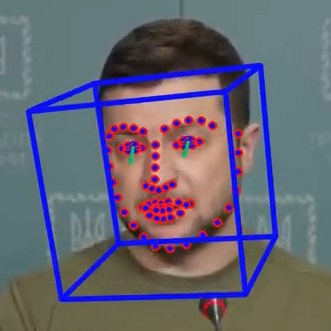} &
        \includegraphics[width=0.165\linewidth]{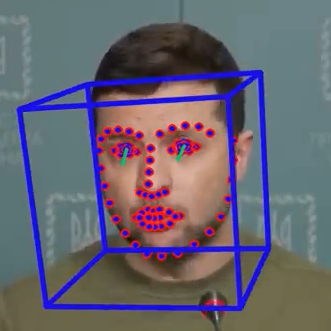} &
        \includegraphics[width=0.165\linewidth]{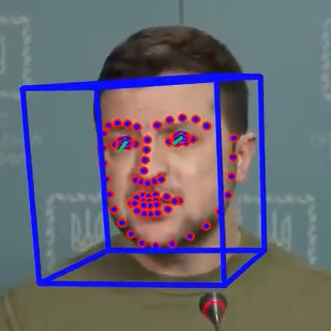} &
        \includegraphics[width=0.165\linewidth]{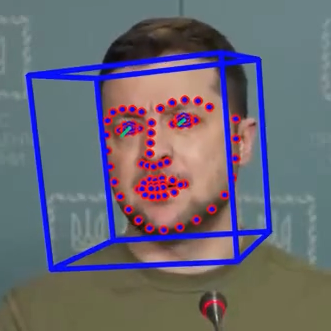} &
        \includegraphics[width=0.165\linewidth]{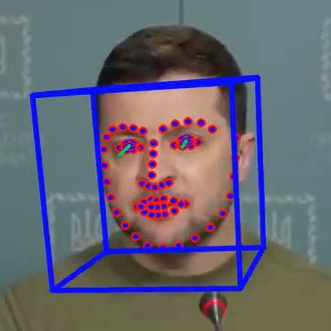} \\ \\
    \end{tabular}
    \begin{tabular}{c}
        \includegraphics[width=0.95\linewidth]{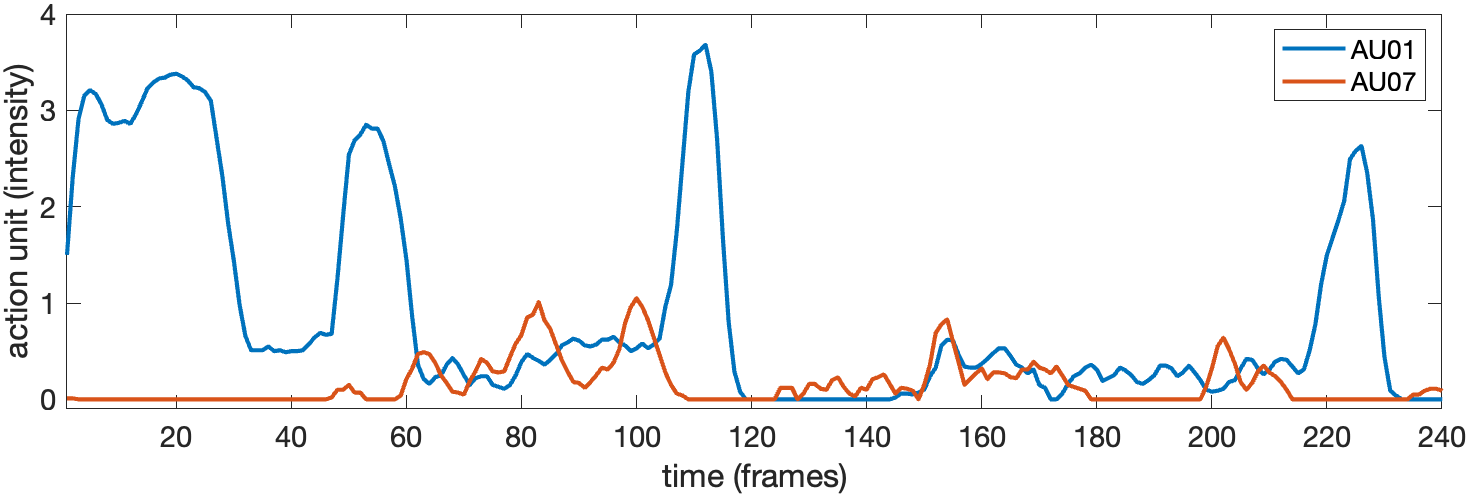}
    \end{tabular}
    \end{center}
    \caption{Shown above are five equally-spaced and cropped frames from a $10$-second video clip annotated with the estimated facial landmarks (red markers) and head pose (blue box). Shown below are two of the $16$ action units as a function of time (inner brow raiser [AU01] and lid tightener [AU07]).}
    \label{fig:facial}
\end{figure}

\subsection{Facial Mannerisms}
\label{subsec:facial}

The identity-based forensic technique of~\citep{agarwal2019protecting} is based on the observation that individuals have distinct speaking styles in terms of facial expressions and head movements. Former President Obama, for example, tends to tilt his head upwards when he smiles, and downwards when he frowns.

Starting with a single video as input, the OpenFace2 toolkit~\citep{baltrusaitis2018openface} extracts facial landmark positions, facial action units, head pose, and eye gaze on a per-frame basis. Facial muscle movement and expression are encoded using facial action units (AU)~\citep{ekman1976measuring}. The OpenFace2 toolkit provides -- on a per-frame basis -- the strength of $17$ different AUs: inner brow raiser (AU01), outer brow raiser (AU02), brow lowerer (AU04), upper lid raiser (AU05), cheek raiser (AU06), lid tightener (AU07), nose wrinkler (AU09), upper lip raiser (AU10), lip corner puller (AU12), dimpler (AU14), lip corner depressor (AU15), chin raiser (AU17), lip stretcher (AU20), lip tightener (AU23), lip part (AU25), jaw drop (AU26), and eye blink (AU45).

The forensic facial model incorporates $16$ AUs (AU45, eye blink, was found not to be distinctive and therefore eliminated from consideration) and four additional features: (1) head rotation about the x-axis; (2) head rotation about the z-axis (as with AU45, head rotation about the y-axis was found not to be distinctive); (3) the horizontal distance between the corners of the mouth (mouth$_h$); and (4) the vertical distance between the lower and upper lip (mouth$_v$), yielding a total of $20$ facial-mannerism features. Shown in Figure~\ref{fig:facial} are several frames of the facial and head tracking and two representative examples of the measured action units across a $10$-second clip. 

These features are combined with gestural mannerisms, described next, to form a person-specific behavioral model.

\begin{figure}[t]
    \begin{center}
    \begin{tabular}{ccccc}
        \includegraphics[width=0.165\linewidth]{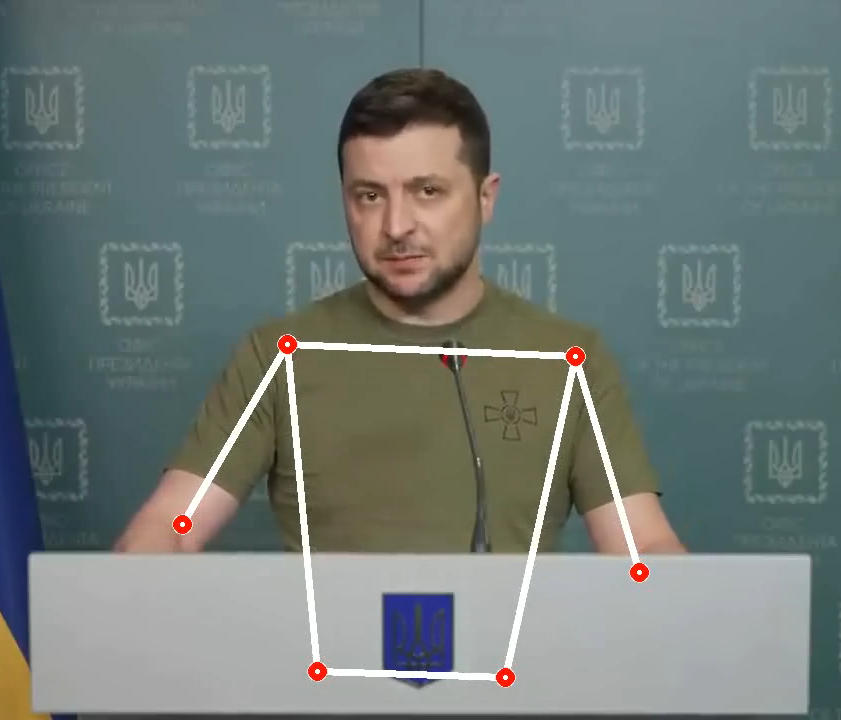} &
        \includegraphics[width=0.165\linewidth]{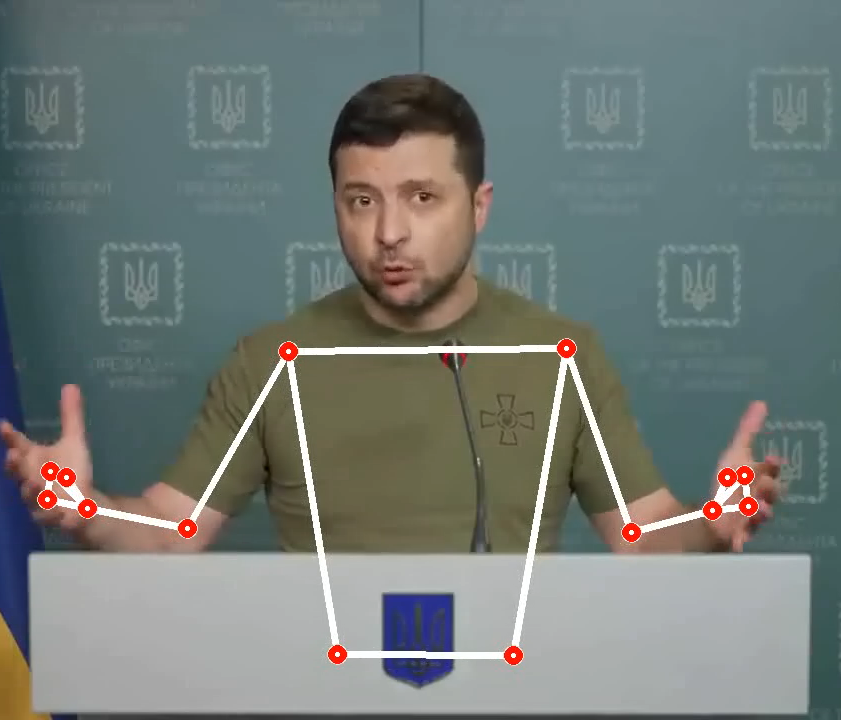} &
        \includegraphics[width=0.165\linewidth]{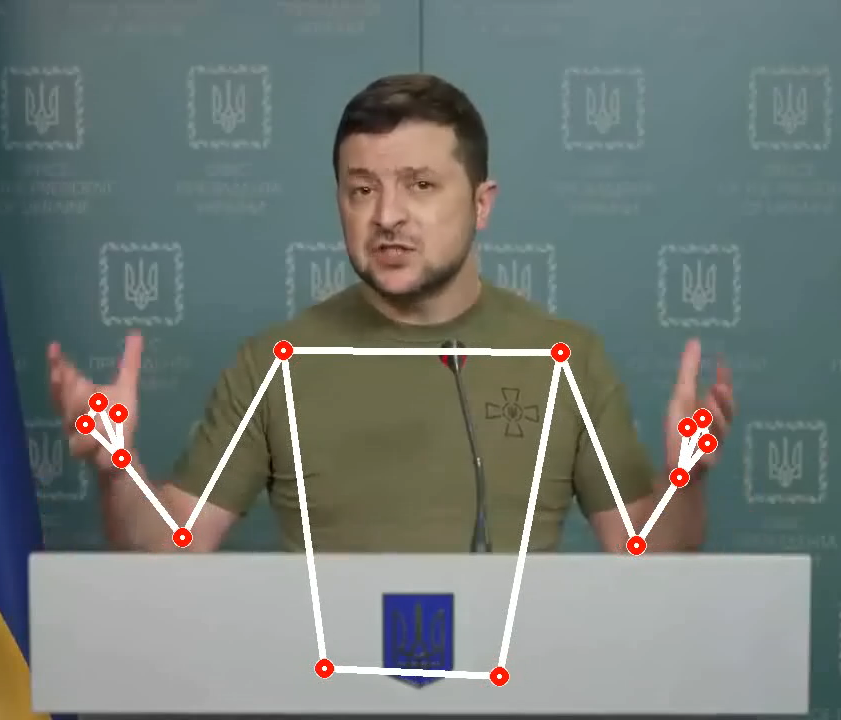} &
        \includegraphics[width=0.165\linewidth]{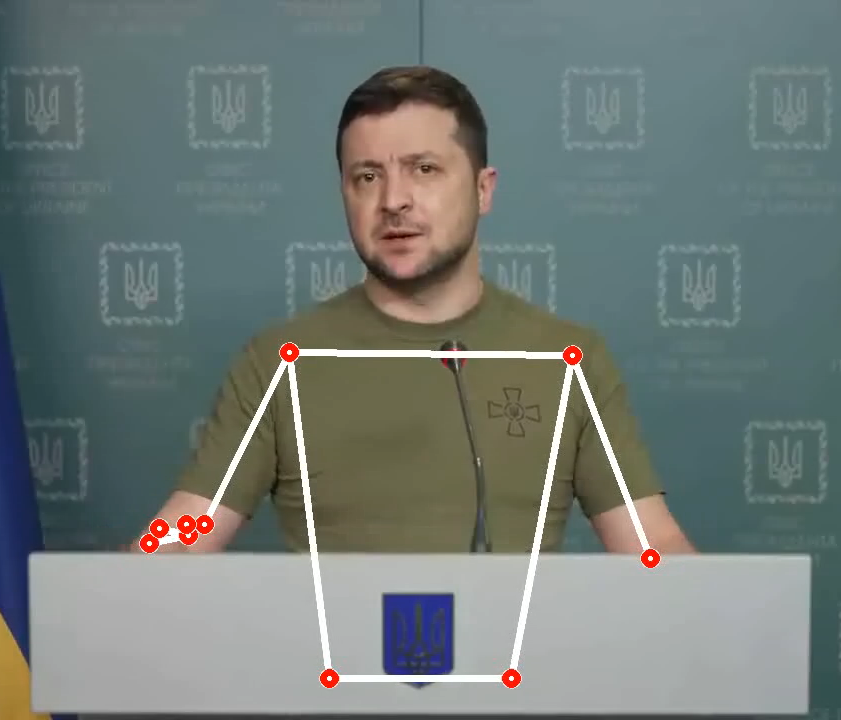} &
        \includegraphics[width=0.165\linewidth]{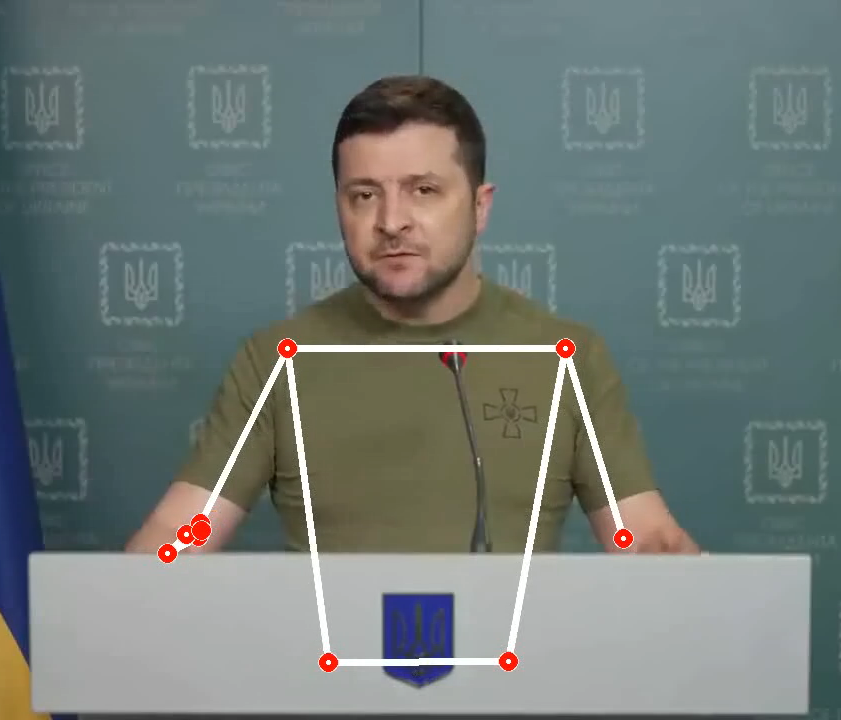} \\ \\
    \end{tabular}
    \begin{tabular}{c}
        \includegraphics[width=0.95\linewidth]{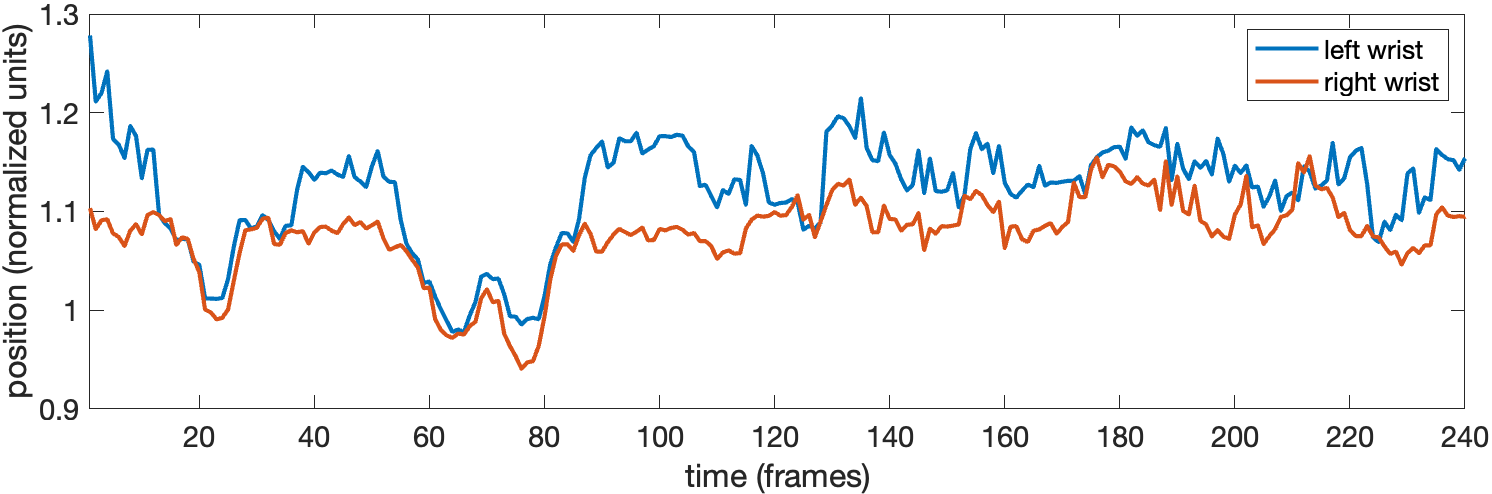} \\
    \end{tabular}
    \end{center}
    \caption{Shown above are five equally-spaced frames from a $10$-second video clip annotated with the estimated gestural tracking. Shown below are two of the $12$ gestural features corresponding to the vertical position of the left and right wrist (the spatial position of the wrists are reported in normalized units relative to a body-centric action plane).}
    \label{fig:gestural}
\end{figure}

\subsection{Gestural Mannerisms}
\label{subsec:gestural}

Across cultures and languages, hand gestures provide additional information not always captured by a speaker's words alone~\citep{church2017gesture}. In addition to distinct gestural patterns found across age~\citep{ozer2017effectsgesture}, sex~\citep{ozccalicskan2010sex}, and culture~\citep{pika2006cross}, recent work also finds that individuals exhibit distinct gestural patterns~\citep{ozer202gesture}. We, therefore, hypothesize that hand gestures, in addition to the facial expressions and head movements described above, will improve our ability to identify an individual's distinct speaking patterns.

Arm and hand position and movement are estimated in each input video frame using  Blazepose~\citep{bazarevsky2020blazepose} from the MediaPipe library~\citep{lugaresi2019mediapipe}. Because we are interested only in the upper body, we consider the image $x$-, $y$-coordinates corresponding to the shoulder, elbow, and wrist of both arms, Figure~\ref{fig:gestural}, yielding a total of $12$ individual measurements. These upper-body coordinates, initially specified relative to the video-frame size, are normalized into a speaker-centric action plane~\citep{bohacek2022sign}. This action plane is a rectangular bounding box centered on the speaker's chest with a width $8 \times$ and height $6 \times$ the measured head height~\citep{De_silva_2008, Bauer_2014}. In this normalized bounding box, the upper left-hand corner is $(0,0)$ and the lower right-hand corner is $(1,1)$. This normalization ensures that the tracked upper-body coordinates can be compared across different speaker locations and sizes. Shown in Figure~\ref{fig:gestural} are several frames of the upper-body tracking and representative examples of the extracted gestural features across a $10$-second clip. 

Whereas the tracked $x,y$ facial features are converted into a higher-level representation in the form of action units, we find that a similar approach with the hand gestures was less effective than simply considering the normalized $x,y$ locations of the tracked shoulders, elbows, and wrists.

\subsection{Behavioral Model}
\label{subsec:behavioralmodel}

Correlations between all pairs of the $20$ facial features and $12$ gestural features are used to capture individualized mannerisms (e.g.,~head tilt and smiling/frowning). A total of $~_{32}C_2 = (32 \times 31)/2 = 496$ correlations are extracted from overlapping $10$-second video clips extracted from an input video in question. 

Trained on authentic video of a person of interest, a novelty detection model in the form of a one-class, non-linear support vector machine (SVM)~\citep{scholkopf2001, scikit-learn} is used to distinguish an individual from imposters and deep fakes. An advantage of this classifier is that it only requires examples of authentic videos. 

The $506$ minutes of Zelenskyy video is partitioned into overlapping (by $5$ seconds) $10$-second video clips, yielding a total of $157,752$ clips. The $110$ minutes of other identities in the World Leaders, FaceForensics++, and Deep-Fake Zelenskyy videos are similarly partitioned, yielding a total of $25,077$ clips.

These clips are randomly partitioned into a $80/20$ training/testing split. The SVM is trained on the $496$ facial- and gestural-feature pairwise correlations. The SVM hyper-parameters, consisting of the Gaussian kernel width ($\gamma$) and outlier percentage ($\nu$), are optimized by performing a grid search over these parameters across the training set. The trained classifier is then evaluated against the hold-out testing set. This entire process is repeated $100$ times with randomized training/testing splits, from which we report average classification accuracy.

Three different classifiers are trained on facial features only, gestural features only, and facial and gestural features combined. The SVM classification threshold for the individual features is selected to yield a $95\%$ training accuracy of correctly classifying real Zelenskyy clips. The classification threshold for the combined features is selected to yield a $99\%$ training accuracy.

\section{Results}
\label{sec:results}

Shown in Table~\ref{tab:results} is the classification accuracy (averaged over $100$ random training/testing data splits) of our behavioral model evaluated against the $10$-second video clips of seven different world-leaders, $28$ distinct identities in the FaceForensics++ dataset~\citep{roessler2019faceforensicspp}, and real and deep-fake versions of Zelenskyy.

We find that the facial features and gestural features alone are insufficient to consistently detect deep-fake version of Zelenskyy (see the last two columns of Table~\ref{tab:results}). The combination of facial and gestural, however, yields significant improvements in detection accuracy. Because deep-fake techniques are -- rightfully -- focused on high-quality facial and audio synthesis, and because of the expected difficulty in synthesizing realistic hands and hand gestures, we posit that the combination of facial and gestural signals will prove reliable for at least a few years.

As compared to the best-performing DFDC model (last row of Table~\ref{tab:results})~\citep{seferbekov2020dfdc}, our model achieves significantly higher classification across all non-Zelenskyy data sets. This comparison, however, is not entirely fair as our behavioral model is trained to detect deep-fake versions of just one identity, whereas the DFDC model is a generic deep-fake detector. On the other hand, our classifier operates on $10$-second video clips whereas the DFDC model has the advantage of operating on the entire video. This comparison does, nevertheless, show the power of identity-specific models.

\begin{table}[t]
\begin{center}
\begin{tabular}{r|SSSSS}
    &                &                  &               & \textbf{Lip-Sync}  & \textbf{In-The-Wild} \\
    & \textbf{World} &                  & \textbf{Real} & \textbf{Deep-Fake} & \textbf{Deep-Fake} \\
      \textbf{Model} & \textbf{Leaders} & \textbf{FF++} & \textbf{Zelenskyy} & \textbf{Zelenskyy} & \textbf{Zelenskyy} \\
    \hline
    facial                          & 91.7 & 91.1 & 94.7 & 17.4 & 83.9 \\
    gestural                        & 77.4 & 95.7 & 95.0 & 12.1 & 33.3 \\
    facial + gestural               & 100.0 & 100.0 & 97.1 & 94.9 & 100.0 \\
    DFDC                            & 73.1 & 84.5 & 93.5 & 13.3 & 1.7
\end{tabular}
\end{center}
\caption{Classification accuracy (reported as percentages) for our behavioral model with facial, gestural, and these two features combined evaluated against seven different world leaders, $28$ identities in the FaceForensics++ dataset and against both real and deep-fake versions of Zelenskyy. By comparison, our model significantly outperforms the best-performing DFDC model~\citep{seferbekov2020dfdc}.}
\label{tab:results}
\end{table}

\subsection{Ablation}

\begin{figure}[t]
    \begin{center}
        \includegraphics[width=0.95\linewidth]{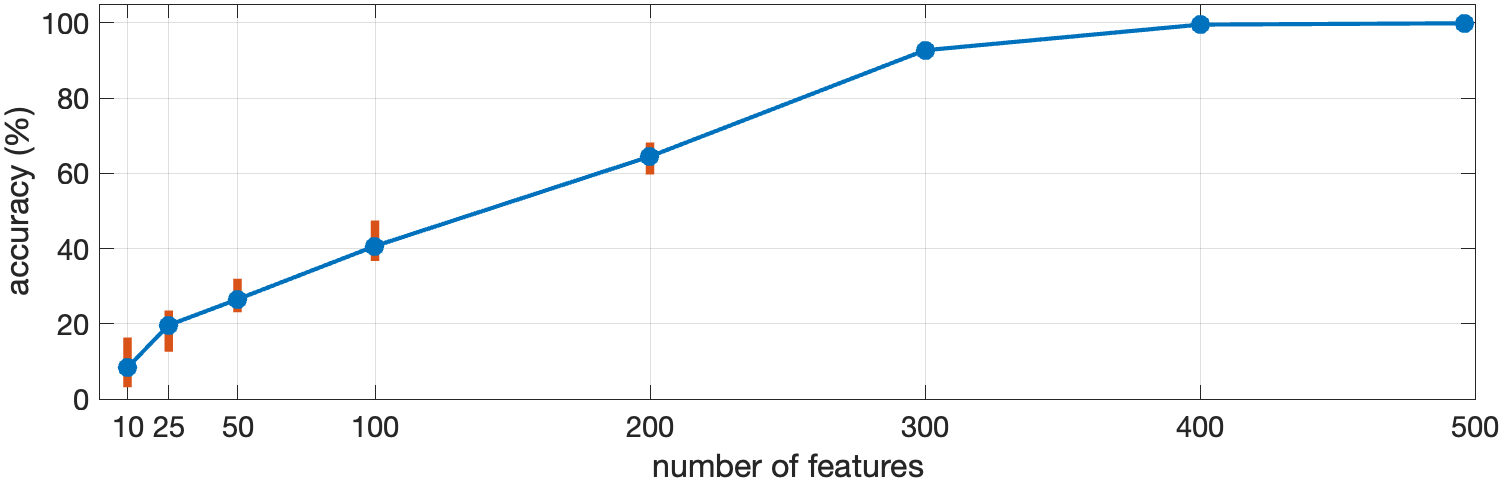}
    \end{center}
    \caption{Each data point corresponds to the median ($50\%$ quantile) accuracy for classifiers trained on between $10$ and $496$ randomly selected facial and gestural features;  the error bars correspond to the $25\%$ and $75\%$ quantile.}
    \label{fig:ablation}
\end{figure}

To determine how many of the $496$ pairwise facial and gestural correlations are needed to achieve the classification accuracy reported in Table~\ref{tab:results}, we trained a series of one-class SVMs on randomly selected subsets -- ranging in size between $10$ and $400$ -- of all facial and gestural features. Shown in Figure~\ref{fig:ablation} is the median accuracy of classifying the identities in the world leaders and deep-fake Zelenskyy videos. In this figure, each data point corresponds to the median accuracy ($50\%$ quantile) from $25$ independent and randomly selected features of each subset size; the error bars correspond to the $25\%$ and $75\%$ quantile. 

With the full set of $496$ facial and gestural features, detection accuracy is $99.88\%$. Detection accuracy grows relatively linearly between feature subsets of size $10$ and $300$ plateauing at $99.52\%$ with $400$ features. Here we see that a significant fraction of the facial and gestural features are, collectively, rich and informative.

To determine which specific facial and gestural features are most discriminative, we next trained $500$ classifiers on random feature subsets of size $10$. The discriminatory power of each feature is computed from the average accuracy of each classifier to which a feature contributed. Across all $500$ classifiers, the detection accuracy on the world leaders and deep-fake Zelenskyy data sets ranges from $44.4\%$ to $4.3\%$. The top $20$ most discriminative correlation features and respective classifier accuracy are:
\begin{center}
    \begin{tabular}{rclc}
    {\bf Feature 1} & & {\bf Feature 2} & {\bf Classifier Accuracy ($\%$)} \\
    \hline
    head-pose-Rx & $\Leftrightarrow$ & right-elbow-y & $44.4$ \\
    head-pose-Rx & $\Leftrightarrow$ & right-wrist-y & $36.9$ \\
    head-pose-Rx & $\Leftrightarrow$ & left-elbow-y & $33.7$ \\
    left-elbow-x & $\Leftrightarrow$ & left-shoulder-x & $32.8$ \\
    head-pose-Rx & $\Leftrightarrow$ & left-shoulder-y & $32.5$ \\
    head-pose-Rx & $\Leftrightarrow$ & lip-vertical & $29.3$ \\
    left-elbow-x & $\Leftrightarrow$ & right-elbow-y & $27.4$ \\
    right-elbow-y & $\Leftrightarrow$ & right-shoulder-y & $27.4$ \\
    right-elbow-x & $\Leftrightarrow$ & right-shoulder-x & $27.3$ \\
    head-pose-Rx & $\Leftrightarrow$ & left-wrist-y & $26.9$ \\
    AU14 & $\Leftrightarrow$ & AU17 & $26.2$ \\
    AU06 & $\Leftrightarrow$ & right-elbow-y & $25.6$\\
    mouth$_v$ & $\Leftrightarrow$ & AU14 & $25.5$ \\
    AU12 & $\Leftrightarrow$ & AU15 & $25.4$ \\
    AU12 & $\Leftrightarrow$ & AU14 & $25.0$ \\
    right-elbow-y & $\Leftrightarrow$ & left-shoulder-y & $25.0$ \\
    pose-Rz & $\Leftrightarrow$ & right-shoulder-x & $24.4$ \\
    mouth$_v$ & $\Leftrightarrow$ & AU15 & $23.7$ \\
    pose-Rx & $\Leftrightarrow$ & right-shoulder-y & $23.0$ \\
    AU06 & $\Leftrightarrow$ & AU14 & $22.7$
 \end{tabular}
\end{center}
where *-Rx and *-Rz correspond to 3-D head rotations, *-x and *-y correspond to the horizontal and vertical image position, and AU* corresponds to specific facial action units (see Section~\ref{subsec:facial}). Here we see that the correlation between head rotation and hand gestural features are the most discriminative, highlighting the importance of the addition of gestural features to the original facial-based model. For President Zelenskyy, in particular, head rotation (as in nodding affirmatively) is highly correlated to his hand movements.

As compared to the median accuracy of $8.4\%$ across random features of subset size $10$ (Figure~\ref{fig:ablation}), these top-ranked features achieve accuracies between three and five times higher. A single classifier trained on the top $10$ and $20$ features, however, only yields a prediction accuracy on the world leaders and deep-fake Zelenskyy data sets of $59.2\%$ and $63.4\%$, providing further evidence that a full set of facial and gestural features are necessary to achieve a high classification accuracy.

\section{Discussion}

Although the term deep fakes first splashed on the screen in 2017, the precursor to what we now call deep fakes dates back two decades. In the seminal video-rewrite work~\cite{bregler1997video}, a video of a person speaking is automatically modified to yield a video of them saying things not found in the original footage. The resulting video quality and resolution were generally lower than today's deep-fake videos, but the results were nevertheless impressive. Some $25$ years later, deep neural networks, GANs, massive data sets, and unlimited compute cycles have led to increasingly more realistic and sophisticated deep-fake videos.

While the democratization of access to techniques for manipulating and synthesizing videos has led to interesting and entertaining applications, they have also given rise to complex ethical and legal question~\cite{chesney2019deep}. In the fog of war, in particular, deep fakes pose a significant threat to our ability to understand and respond to rapidly evolving events. 

While our approach to protecting a single individual -- Ukrainian President Zelenskyy -- does not address the broader issue of deep fakes, it does bring some level of digital protection to the arguably most important Ukrainian voice at this time of war.

\section*{Acknowledgement}

We are grateful to Zoltan Kovacs, Muhammad Shahzaib Aslam, and the rest of the Colossyan team for creating the Zelenskyy lip-sync deep fakes.

\bibliographystyle{unsrtnat}
\bibliography{refs}

\begin{thebibliography}{33}
\providecommand{\natexlab}[1]{#1}
\providecommand{\url}[1]{\texttt{#1}}
\expandafter\ifx\csname urlstyle\endcsname\relax
  \providecommand{\doi}[1]{doi: #1}\else
  \providecommand{\doi}{doi: \begingroup \urlstyle{rm}\Url}\fi

\bibitem[Allyn(2022)]{allyn2022deepfake}
Bobby Allyn.
\newblock Deepfake video of {Z}elenskyy could be `tip of the iceberg' in info
  war, experts warn.
\newblock
  \url{https://www.npr.org/2022/03/16/1087062648/deepfake-video-zelenskyy-experts-war-manipulation-ukraine-russia},
  2022.

\bibitem[Groh et~al.(2022)Groh, Epstein, Firestone, and
  Picard]{groh2022deepfake}
Matthew Groh, Ziv Epstein, Chaz Firestone, and Rosalind Picard.
\newblock Deepfake detection by human crowds, machines, and machine-informed
  crowds.
\newblock \emph{Proceedings of the National Academy of Sciences}, 119\penalty0
  (1), 2022.

\bibitem[Ferrer(2020)]{dfdc2020}
Cristian~Canton Ferrer.
\newblock Deepfake detection challenge results: {A}n open initiative to advance
  {AI}.
\newblock
  \url{https://ai.facebook.com/blog/deepfake-detection-challenge-results-an-open-initiative-to-advance-ai},
  2020.

\bibitem[Zhou et~al.(2017)Zhou, Han, Morariu, and Davis]{zhou2017twostream}
Peng Zhou, Xintong Han, Vlad~I. Morariu, and Larry~S. Davis.
\newblock Two-stream neural networks for tampered face detection.
\newblock In \emph{International Conference on Computer Vision and Pattern
  Recognition}, 2017.

\bibitem[Afchar et~al.(2018)Afchar, Nozick, Yamagishi, and
  Echizen]{afchar2018mesonet}
Darius Afchar, Vincent Nozick, Junichi Yamagishi, and Isao Echizen.
\newblock {MesoNet}: A compact facial video forgery detection network.
\newblock In \emph{IEEE International Workshop on Information Forensics and
  Security}, 2018.

\bibitem[Li et~al.(2019)Li, Bao, Zhang, Yang, Chen, Wen, and Guo]{li2019xray}
Lingzhi Li, Jianmin Bao, Ting Zhang, Hao Yang, Dong Chen, Fang Wen, and Baining
  Guo.
\newblock Face {X-ray} for more general face forgery detection.
\newblock arXiv:1912.13458, 2019.

\bibitem[Li et~al.(2018)Li, Chang, and Lyu]{li2018ictu}
Yuezun Li, Ming-Ching Chang, and Siwei Lyu.
\newblock In ictu oculi: {E}xposing {AI} created fake videos by detecting eye
  blinking.
\newblock In \emph{International Workshop on Information Forensics and
  Security}, pages 1--7, 2018.

\bibitem[Agarwal and Farid(2021)]{agarwal2021detecting}
Shruti Agarwal and Hany Farid.
\newblock Detecting deep-fake videos from aural and oral dynamics.
\newblock In \emph{CVPR Workshop on Media Forensics}, pages 981--989, 2021.

\bibitem[Agarwal et~al.(2020{\natexlab{a}})Agarwal, Farid, Fried, and
  Agrawala]{agarwal2020phoneme}
Shruti Agarwal, Hany Farid, Ohad Fried, and Maneesh Agrawala.
\newblock Detecting deep-fake videos from phoneme-viseme mismatches.
\newblock In \emph{CVPR Workshop on Media Forensics}, pages 660--661,
  2020{\natexlab{a}}.

\bibitem[Agarwal et~al.(2019)Agarwal, Farid, Gu, He, Nagano, and
  Li]{agarwal2019protecting}
Shruti Agarwal, Hany Farid, Yuming Gu, Mingming He, Koki Nagano, and Hao Li.
\newblock Protecting world leaders against deep fakes.
\newblock In \emph{CVPR Workshop on Media Forensics}, volume~1, 2019.

\bibitem[Agarwal et~al.(2020{\natexlab{b}})Agarwal, Farid, El-Gaaly, and
  Lim]{agarwal2020appearance}
Shruti Agarwal, Hany Farid, Tarek El-Gaaly, and Ser-Nam Lim.
\newblock Detecting deep-fake videos from appearance and behavior.
\newblock In \emph{International Workshop on Information Forensics and
  Security}, pages 1--6, 2020{\natexlab{b}}.

\bibitem[Agarwal et~al.(2021)Agarwal, Hu, Ng, Darrell, Li, and
  Rohrbach]{agarwal2021watch}
Shruti Agarwal, Liwen Hu, Evonne Ng, Trevor Darrell, Hao Li, and Anna Rohrbach.
\newblock Watch those words: {V}ideo falsification detection using
  word-conditioned facial motion.
\newblock arXiv:2112.10936, 2021.

\bibitem[Cozzolino et~al.(2021)Cozzolino, R{\"o}ssler, Thies, Nie{\ss}ner, and
  Verdoliva]{cozzolino2021id}
Davide Cozzolino, Andreas R{\"o}ssler, Justus Thies, Matthias Nie{\ss}ner, and
  Luisa Verdoliva.
\newblock {ID}-reveal: {I}dentity-aware deepfake video detection.
\newblock In \emph{International Conference on Computer Vision}, pages
  15108--15117, 2021.

\bibitem[Carlini and Farid(2020)]{carlini2020evading}
Nicholas Carlini and Hany Farid.
\newblock Evading deepfake-image detectors with white-and black-box attacks.
\newblock In \emph{CVPR Workshop on Media Forensics}, pages 658--659, 2020.

\bibitem[Barni et~al.(2018)Barni, Stamm, and Tondi]{barni2018adversarial}
Mauro Barni, Matthew~C Stamm, and Benedetta Tondi.
\newblock Adversarial multimedia forensics: {O}verview and challenges ahead.
\newblock In \emph{European Signal Processing Conference}, pages 962--966,
  2018.

\bibitem[R\"ossler et~al.(2019)R\"ossler, Cozzolino, Verdoliva, Riess, Thies,
  and Nie{\ss}ner]{roessler2019faceforensicspp}
Andreas R\"ossler, Davide Cozzolino, Luisa Verdoliva, Christian Riess, Justus
  Thies, and Matthias Nie{\ss}ner.
\newblock Face{F}orensics++: {L}earning to detect manipulated facial images.
\newblock In \emph{International Conference on Computer Vision and Pattern
  Recognition}, 2019.

\bibitem[Baltrusaitis et~al.(2018)Baltrusaitis, Zadeh, Lim, and
  Morency]{baltrusaitis2018openface}
Tadas Baltrusaitis, Amir Zadeh, Yao~Chong Lim, and Louis-Philippe Morency.
\newblock Openface 2.0: {F}acial behavior analysis toolkit.
\newblock In \emph{IEEE International Conference on Automatic Face \& Gesture
  Recognition}, pages 59--66, 2018.

\bibitem[Ekman and Friesen(1976)]{ekman1976measuring}
Paul Ekman and Wallace~V Friesen.
\newblock Measuring facial movement.
\newblock \emph{Environmental Psychology and Nonverbal Behavior}, 1\penalty0
  (1):\penalty0 56--75, 1976.

\bibitem[Church et~al.(2017)Church, Alibali, and Kelly]{church2017gesture}
R.~Breckinridge Church, Marha~W. Alibali, and Spencer~D. Kelly.
\newblock \emph{Why Gesture?: How the hands function in speaking, thinking and
  communicating}, volume~7.
\newblock John Benjamins Publishing Company, 2017.

\bibitem[Özer D. et~al.(2017)Özer D., M., Özer E.~E., K., A., and
  T]{ozer2017effectsgesture}
Özer D., Tansan M., Özer E.~E., Malykhina K., Chatterjee A., and Göksun T.
\newblock The effects of gesture restriction on spatial language in young and
  elderly adults.
\newblock In \emph{38th Annual Conference of the Cognitive Science Society},
  pages 1471–--1476, 2017.

\bibitem[{\"O}z{\c{c}}al{\i}{\c{s}}kan and
  Goldin-Meadow(2010)]{ozccalicskan2010sex}
{\c{S}}eyda {\"O}z{\c{c}}al{\i}{\c{s}}kan and Susan Goldin-Meadow.
\newblock Sex differences in language first appear in gesture.
\newblock \emph{Developmental Science}, 13\penalty0 (5):\penalty0 752--760,
  2010.

\bibitem[Pika et~al.(2006)Pika, Nicoladis, and Marentette]{pika2006cross}
Simone Pika, Elena Nicoladis, and Paula~F Marentette.
\newblock A cross-cultural study on the use of gestures: Evidence for
  cross-linguistic transfer?
\newblock \emph{Bilingualism: Language and Cognition}, 9\penalty0 (3):\penalty0
  319--327, 2006.

\bibitem[Özer D and T.(2020)]{ozer202gesture}
Özer D and Göksun T.
\newblock Gesture use and processing: A review on individual differences in
  cognitive resources.
\newblock \emph{Frontiers in Psycholology}, 2020.

\bibitem[Bazarevsky et~al.(2020)Bazarevsky, Grishchenko, Raveendran, Zhu,
  Zhang, and Grundmann]{bazarevsky2020blazepose}
Valentin Bazarevsky, Ivan Grishchenko, Karthik Raveendran, Tyler Zhu, Fan
  Zhang, and Matthias Grundmann.
\newblock Blazepose: On-device real-time body pose tracking.
\newblock arXiv:2006.10204, 2020.

\bibitem[Lugaresi et~al.(2019)Lugaresi, Tang, Nash, McClanahan, Uboweja, Hays,
  Zhang, Chang, Yong, Lee, et~al.]{lugaresi2019mediapipe}
Camillo Lugaresi, Jiuqiang Tang, Hadon Nash, Chris McClanahan, Esha Uboweja,
  Michael Hays, Fan Zhang, Chuo-Ling Chang, Ming~Guang Yong, Juhyun Lee, et~al.
\newblock Mediapipe: A framework for building perception pipelines.
\newblock arXiv:1906.08172, 2019.

\bibitem[Boháček and Hrúz(2022)]{bohacek2022sign}
Matyáš Boháček and Marek Hrúz.
\newblock Sign pose-based transformer for word-level sign language recognition.
\newblock In \emph{IEEE/CVF Winter Conference on Applications of Computer
  Vision Workshops}, pages 182--191, 2022.

\bibitem[De~Silva(2008)]{De_silva_2008}
Liyanage De~Silva.
\newblock Audiovisual sensing of human movements for home-care and security in
  a smart environment.
\newblock \emph{International Journal On Smart Sensing and Intelligent
  Systems}, 1, 2008.

\bibitem[Bauer(2014)]{Bauer_2014}
Anastasia Bauer.
\newblock \emph{The Use of Signing Space in a Shared Sign Language of
  Australia}.
\newblock De Gruyter, 1 edition, 2014.

\bibitem[Sch\"{o}lkopf et~al.()Sch\"{o}lkopf, Platt, Shawe-Taylor, Smola, and
  Williamson]{scholkopf2001}
Bernhard Sch\"{o}lkopf, John~C. Platt, John~C. Shawe-Taylor, Alex~J. Smola, and
  Robert~C. Williamson.
\newblock Estimating the support of a high-dimensional distribution.
\newblock \emph{Neural Computation}, 13\penalty0 (7):\penalty0 1443--1471.

\bibitem[Pedregosa et~al.(2011)Pedregosa, Varoquaux, Gramfort, Michel, Thirion,
  Grisel, Blondel, Prettenhofer, Weiss, Dubourg, Vanderplas, Passos,
  Cournapeau, Brucher, Perrot, and Duchesnay]{scikit-learn}
F.~Pedregosa, G.~Varoquaux, A.~Gramfort, V.~Michel, B.~Thirion, O.~Grisel,
  M.~Blondel, P.~Prettenhofer, R.~Weiss, V.~Dubourg, J.~Vanderplas, A.~Passos,
  D.~Cournapeau, M.~Brucher, M.~Perrot, and E.~Duchesnay.
\newblock Scikit-learn: {M}achine learning in {P}ython.
\newblock \emph{Journal of Machine Learning Research}, 12:\penalty0 2825--2830,
  2011.

\bibitem[Seferbekov(2020)]{seferbekov2020dfdc}
Selim Seferbekov.
\newblock {DFDC} deepfake challenge solution.
\newblock \url{https://github.com/selimsef/dfdc_deepfake_challenge}, 2020.

\bibitem[Bregler et~al.(1997)Bregler, Covell, and Slaney]{bregler1997video}
Christoph Bregler, Michele Covell, and Malcolm Slaney.
\newblock Video rewrite: Driving visual speech with audio.
\newblock In \emph{24th Annual Conference on Computer Graphics and Interactive
  Techniques}, pages 353--360, 1997.

\bibitem[Chesney and Citron(2019)]{chesney2019deep}
Bobby Chesney and Danielle Citron.
\newblock Deep fakes: A looming challenge for privacy, democracy, and national
  security.
\newblock \emph{California Law Review}, 107:\penalty0 1753, 2019.

\end{thebibliography}

\end{document}